\documentclass[letter, 10 pt, conference]{ieeeconf}  

\IEEEoverridecommandlockouts                              

\usepackage{graphics} 
\usepackage{epsfig} 
\usepackage[linesnumbered,ruled]{algorithm2e}
\usepackage{subcaption}
\usepackage{float}
\usepackage{graphicx}
\usepackage{bm}
\usepackage{times}
\usepackage{balance}
\usepackage{amsmath}
\usepackage{breqn}
\usepackage{cite}
\usepackage{threeparttable}
\usepackage{multirow}
\usepackage{booktabs}
\usepackage{bigstrut}
\usepackage{amsmath}
\usepackage{makecell}
\usepackage{color,soul}

\DeclareMathOperator*{\argmin}{arg\,min}

\newcommand*{\argminl}{\argmin\limits}

\title{\LARGE \bf
    Robust Event Detection based on Spatio-Temporal Latent Action Unit using Skeletal Information 
}

\author{Hao Xing$^{1}$ Yuxuan Xue$^{1}$ Mingchuan Zhou$^2$ and Darius Burschka$^{1}$
	\thanks{${^{1}}$Authors  are with Machine Vision and Perception Group, Department of Computer Science,
		Technical University of Munich, Arcisstraße $21$, $80333$ Munich, Germany
		{\tt\small hao.xing@tum.de}, {\tt\small yuxuan.xue@tum.de}, {\tt\small burschka@cs.tum.edu}}%
	\thanks{$^{2}$Author is with Chair for Computer Aided Medical Procedures and Augmented Reality, Technical University of Munich
		{\tt\small mingchuan.zhou@in.tum.de}}%
}%

\begin{document}

	\maketitle
	\thispagestyle{empty}
	\pagestyle{empty}

	\begin{abstract}

		This paper proposes a novel dictionary learning approach to detect event anomalities using skeletal information extracted from RGBD video. The event action is represented as several latent action atoms and composed of latent spatial and temporal attributes. We aim to construct a network able to learn from few examples and also rules defined by the user. The skeleton frames are clustered by an initial K-means method. Each skeleton frame is assigned with a varying weight parameter and fed into our Gradual Online Dictionary Learning (GODL) algorithm. During the training process, outlier frames will be gradually filtered by reducing the weight that is inversely proportional to a cost. To strictly distinguish the event action from similar actions and robustly acquire its action units, we build a latent unit temporal structure for each sub-action.  
		
		We validate the method at the example of fall event detection on NTU RGB+D dataset, because it provides a benchmark available for comparison.
		We present the experimental validation of the achieved accuracy, recall, and precision. Our approach achieves the best performance in precision and accuracy of human fall event detection, compared with other existing dictionary learning methods. Our method remains the highest accuracy and the lowest variance, with increasing noise ratio. 
	\end{abstract}

	\section{INTRODUCTION}

    An important aspect in human-robot interaction is the recognition of trigger events that require a response on the robot side \cite{laptev2007retrieving, turaga2008machine, fan2009recognition}. Such an event is usual an unexpected action or motion of the human subject. These events may trigger additional learning of new motions or an emergency response in case of accidents. In this paper, we address a common event detection of humans falling down due to tripping or health conditions.

	
	
	Since the most part of human body can be viewed as an articulated system with rigid bones connected by joints, human action can be expressed as the movement of skeleton \cite{lie2019fully}.
	Most existing skeleton based event detection methods can be generally categorized into two categories: 2D skeleton-based \cite{lie2018human, avola20192, zheng2019fall} and 3D skeleton-based \cite{min2018support, wu2019skeleton, zhang2016automatic}. 
    Compared to 2D skeleton-based methods, the 3D skeleton has more extensive spatial information at the cost of higher time-consuming and manual labeling requirements. Most existing research methods still have an ill-posed and inverse problem that extracts 3D skeleton from monocular images \cite{zheng2020deep}.
    
    The emergence of Microsoft Kinect \cite{kinect}, and RealSense \cite{realsense} cameras made multidimensional observation of human events feasible without high processing loads on the system.
    However, the noise of the depth measurement in these cameras has a significant influence on event detection. To solve the problem, we applied a gradual filtering processing on skeleton sequences extracted from RGB images using a lightweight Deep Learning toolbox with aligned depth information.
    


	In addition to detecting the event action, learning and establishing structure representation of the action is also essential and challenging. Different actions may have the same start, end position, and similar pose transformation and rotation, such as lying down and fall down. However, their latent temporal feature is totally different. Modeling latent spatio-temporal structures of actions is one of the most widely-used techniques for action recognition, and representation \cite{rabiner1989tutorial, wang2010hidden, tang2012learning}. A latent spatio-temporal structure has two parts: action unit with spatial information and temporal model. The action units are the sequence and constituent elements of action. The temporal feature defines the length of step from the previous state to the next state \cite{qi2018learning}. For the fall-down event, the temporal feature is the sharp height change of skeleton \cite{ma2014depth}.
	
	For the latent action unit extraction, Sparse Coding Dictionary (SCD) is a well-known approach \cite{chiang2013multi, ben2018coding, mairal2010online}, which approximate a given video sequence $\mathbf{Y}$ by the manipulation of a low-rank dictionary $\mathbf{D}$ and its coefficient matrix $\mathbf{X}$. Online Dictionary Learning is one of the most successful SCD methods and is widely used in the action recognition area. Because that fall event detection is just one extreme case of action recognition, we consider the ODL algorithm in this work as a baseline method. Its cost can be expressed in the least squares problem with regularizer as
	\begin{equation}\label{eq:1}
	\min_{\forall \mathbf{D}_i \in \mathcal{D}, \forall \mathbf{X}_i \in \mathcal{X}} \sum_{i=1}^{N}\frac{1}{2}\parallel\mathbf{Y}_{i}-\mathbf{D}_{i}\mathbf{X}_{i} \parallel_{F}^{2} + \lambda\parallel\mathbf{X}_{i}\parallel
	\end{equation}
	where $F$ means Frobenius norm, $N$ is the number of action unit and $\lambda$ is the regularization parameter. Unfortunately, in the presence of outliers, Eq~(\ref{eq:1}) provides a poor estimation for $\mathbf{D}$ and $\mathbf{X}$ \cite{yang2020graduated}. The performance is worse for the 3D skeleton-based human fall event detection because the 3D skeleton has more outlier sources, such as skeleton estimation and depth measurement. 
	
	In this paper, an attempt to improve event detection latency and temporal resolution is presented and performed at the example of fall detection. We separate the fall event into five latent action atoms "\textit{standing}", "\textit{bending knee}", "\textit{opening arm}", "\textit{Knee landing}" and "\textit{arm supporting}". 
	
	Overall, the technical contributions of the paper are:
	\begin{itemize}
		\item    We propose a novel Gradual Online Dictionary Learning method that uses Graduated Non-convexity (GNC) with Geman McClure (GM) cost function to decrease outlier weight during training.
		\item We demonstrate that our approach can robustly extract action unit and detect fall events with training data with a different ratio of outlier. 
		\item We compare our results with other good dictionary learning approaches on the NTU RGB+D dataset \cite{Shahroudy_2016_NTURGBD} and achieve the best performance in the aspect of precision and accuracy.
	\end{itemize}
	
	The rest of the paper is organized as follows: in section \ref{sec:2}, we briefly review existing approaches of the latent action unit and the Sparse Coding Dictionary. Section \ref{sec:3} introduces Gradual Sparse Coding Dictionary. Section \ref{sec:4} reports experimental results and discussions. Section \ref{sec:5} concludes the paper.
	
	\section{RELATED WORK}
	\label{sec:2}
	We review the previous works from three primary related streams of the research area: fall-down event detection, Spatio-temporal latent action unit extraction, and global minimization with a robust cost. 
	\subsection{Fall-Down Event Detection}
	With the rapid development of motion capture technologies, e.g., single RGB camera systems \cite{rougier2011robust, de2017home, huang2018video, mirmahboub2012automatic, tra2013human}, fall event detection has recently received growing attention because of its importance in the health-care area. 
	
	For 3D event detection, RGBD cameras, e.g., Microsoft Kinect and Intel RealSense, provide a significant advantage over standard  cameras\cite{wei2019learning}. Nghiem et al. \cite{nghiem2012head} proposed a method to detect falling down, based on the speed of head and body centroid and their distance to the ground.
	Stone et al. \cite{stone2014fall} used Microsoft Kinect to obtain person's vertical state from depth image frames based on ground segmentation. Fall is detected by analyzing the velocity from the initial state until the human is on the ground. 
	In contrast with using depth images directly, Volkhardt et al. \cite{volkhardt2013fallen} segmented and classified the point cloud from depth images to detect fall events.
	
	Since depth-based methods are sensitive to the error of shape and depth \cite{wei2019learning}, many researchers prefer 3D skeleton-based methods. Tran \cite{le2014analysis} computed three states (distance, angle, velocity) from Kinect's 3D skeleton and applied support vector machine (SVM) to classify falling down action. Kong et al. \cite{kong2018privacy} applied Fast Fourier Transform (FFT) to classify the 3D fall event skeleton dataset. However, the 3D skeleton estimation using a monocular camera is an ill-posed and inverse problem \cite{zheng2020deep}.
	
	\subsection{Spatio-Temporal Latent Action Unit Extraction}

    Based on sparse coding and dictionary learning method, falling down action can be represented as a linear combination of dictionary elements (latent action units). 
	After Mairal et al. \cite{mairal2010online} proposed an Online Dictionary Learning algorithm. It has attracted a lot of attention because of its robustness \cite{chiang2013multi, ferrari2017dictionary, qi2018learning, wilson2014dictionary}. Ramirez et al. \cite{ramirez2010classification} proposed a classic Dictionary Learning method with Structured Incoherence (DLSI) considering the incoherence between different dictionaries as part of the cost, which could have shared atoms between dictionary.
	In against sharing dictionary, Yang et al. \cite{yang2011fisher} presented Fisher Discrimination Dictionary Learning (FDDL) using both the discriminative information in the reconstruction error and sparse coding coefficients to maximize the distance between dictionary. In other words, one training data should only be approximated by the dictionary generated from its cluster.
	Kong et al. \cite{kong2012dictionary} separated the dictionary into Particularity and Commonality and proposed a novel dictionary learning method COPAR. 
	With the similar idea, Tiep et al. \cite{vu2016learning} developed Low-Rank Shared Dictionary Learning (LRSDL) that extract a bias matrix for all dictionary based on FDDL. However, its performance is limited for action recognition because each action unit should have a different action space. The results are discussed in the evaluation chapter.

	
	Recently, spatio-temporal deep convolutional networks \cite{plizzari2020spatial,wen2019graph, yan2018spatial,chen2020afnet, li2020spatio} have been widely applied for action recognition. The common principle of these works is that using several continuous frames generate temporal information around feature joints. 
	However, the size of temporal block is a tricky problem among different actions. Besides that, some events have a strict sequence, such as fall down starts from standing (sitting) and ends on the ground. Most of the deep learning networks cannot identify the sequence by summing all temporal blocks note.
	
	
	\subsection{Global Minimization with Robust Cost}
	Global minimization of ODL is NP-hard with respect to both outliers and chosen of regularization parameters.
	RANSAC \cite{fischler1981random} is a widely used approach but does not guarantee optimality and its calculation time increases exponentially with the outlier rate \cite{yang2020graduated}. The Graduated Non-convexity has also been successfully applied in Computer Vision tasks to optimize robust costs \cite{nielsen1995surface}\cite{rangarajan1990generalized}. However, with a lack of non-minimal solvers, GNC is limited to be used for spatial perception.
	Zhou et al. \cite{zhou2016fast} proposed a fast global registration method, which combines the least square cost with weight function by Black-Rangarajan duality. Yang et al. \cite{yang2020graduated} applied this method to 3D point cloud registration and pose graph estimation. 
	
	Inspired by the successful works mentioned above, we propose a novel dictionary learning method to robustly extract spatial and temporal latent action units under noised by depth image and uncertainty of 2D human pose estimation. 
	
	\begin{figure*}[t]
	\vspace{0.13cm}
	\centering
		\includegraphics[width=1\textwidth]{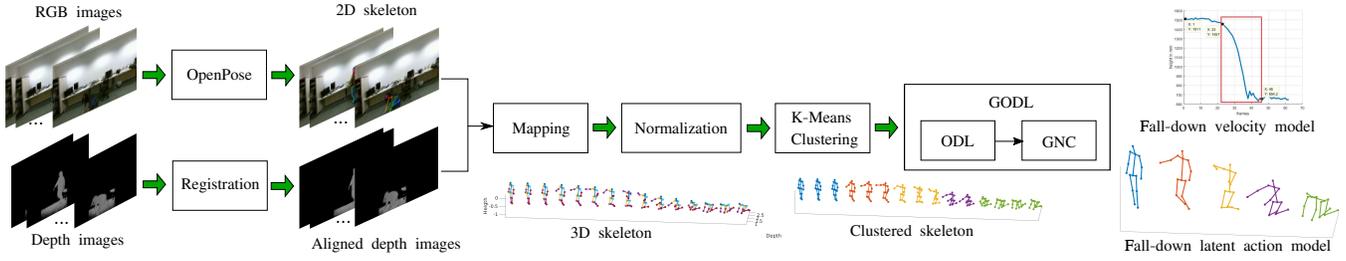}
		\caption{Overview of human fall event detection training process.}
		\label{fig:system}
	\end{figure*}
	\section{METHOD}
	\label{sec:3}
	In this section, we first briefly introduce the setting of GODL and then present our framework.
	
	\subsection{Task Definition}
	Formally, let $\mathbf{Y} = \{\bar{y}_{1},\dots,\bar{y}_{t}\}$ denote a fall-down 3D pose sequence and $\bar{y}_{j}$ is the $j$-th column vector of skeleton joints. 
	We assume that the sequence $\mathbf{Y}$ is segmented into $N$ sub-sequences $\{\mathbf{Y}_{1},\dots,\mathbf{Y}_{N}\}$ and each sub-sequence corresponds to an action unit $\mathbf{D}_{i}=\{\bar{d}_{1},\dots,\bar{d}_{k}\}$. Then the dictionary can be expressed as $\mathbf{D} = \{\mathbf{D}_{1},\dots,\mathbf{D}_{N}\}$ and their coefficient matrix is defined as $\mathbf{X} = \{\mathbf{X}_{1},\dots,\mathbf{X}_{N}\}$. 
	

	\subsection{Prepossessing of data}
	

	An overview of the fall event detection training process is shown in Fig~\ref{fig:system}. RGB images are fed into OpenPose to get 2D skeleton joints. At the same time, depth frames are aligned with RGB images. 3D skeleton joints are obtained by projecting pixel position to 3D space along with aligned depth value. In order to compensate the fact that human could fall down from different positions in image coordinate, a normalization function is applied to keep skeletons in the same magnitude and the ratio for each direction: $x\in[0,1], y\in[0, y_{max}/(x_{max} -x_{min})], z \in [0, z_{max}/(x_{max}-x_{min})]$. In order to balance the influence of spatial and temporal information, we use a weight parameter $w_{{s/t}}$ that is $0.1$ and defined as $w_{{s/t}} = p/v$ in the paper. A K-means based clustering method segments a sequence into $N$ clusters. 
	
	\subsection{Train phase: Gradual Online Dictionary Learning}
	For each sub-sequence, we apply GODL to iteratively update the coefficient matrix $\mathbf{X}_i$ and its action unit matrix $\mathbf{D}_i$, until the cost converges or the maximum iteration number is reached. The general framework for GODL is described in Algorithm~\ref{tab:godl}. The main idea is to automatically enable the iteration process to automatically filter outliers and ensure that the latent action units are learned from inliers.
	\begin{algorithm}[t]
		\SetKwInOut{Input}{Input}
		\SetKwInOut{Output}{Output}
		
		\Input{Fall-down 3D skeleton sequence $\mathbf{Y}$}
		\Output{Dictionary matrix $\mathbf{D}$ and coefficient matrix $\mathbf{X}$}
		
		\SetKwRepeat{Do}{do}{while}
		\While{$i<N$}{
			\textbf{Initialization:} $\bar{w}_i^{T(0)}=\mathbf{\bar{1}}^T$, $\mu_0 = 2*e^2_{i,max}/c^2, \mathbf{D}_i^{(0)}, \mathbf{X}_i^{(0)}$ \;
			
			\While{$\mu \geq 1$}{
				Filter outlier from $\mathbf{Y}_i$: ($\cdot$ is column dot-production)\\
				$\mathbf{\hat{Y}_i} = \bar{w}_{i}^{T}\cdot\mathbf{Y}_i\;\; \text{and}\;\;  \bar{\lambda}_w = \lambda\bar{w}_i^{T}$\;
				\Repeat{end of ODL iteration or reach convergence}{
					Update $\mathbf{\hat{X}}_{i}^{(k)}$ with fixed $\mathbf{D}_{i}^{(k-1)}$\;
					Update $\mathbf{D}_{i}^{(k)}$ with fixed $\mathbf{\hat{X}}_{i}^{(k)}$\;
				}
				Update weight vector:\\
				\SetKw{KwBy}{by}
				\For{$j\gets t_i^1$ \KwTo $t_i^{\mathrm{end}}$}{
					$w_{i,j}^{(k)}=\argminl_{w_i \in [0,1]} 	\mathcal{O}_{i,j} + \Phi_{g_{\mu}}$ \\
				}
				Update $\mu = \mu/1.4$ \;
			}
			
		}
		\caption{Gradual Online Dictionary Learning}
		\label{tab:godl}
	\end{algorithm}

	Graduated non-convexity is a popular method for optimizing general non-convexity cost functions like Geman McClure (GM) function. The following equation shows the GM function:
	\begin{equation}
	g_{\mu}(e) \dot{=} \frac{\mu c^2e^2}{\mu c^2+e^2}
	\end{equation}
	where $c^2$ is a given constant that is the maximum accepted error of inliers, $\mu$ determines the shape of GM function and $e^2$ is Frobenius norm of error between training sequence $\bar{y}_{i,j}$ and approximation model $\mathbf{D}_{i}\bar{x}^{T}_{i,j}$ as follow:
	\begin{equation}
	\begin{split}\label{eq:error}
	e_{i,j}^2 &= \parallel \bar{y}_{i,j}-\mathbf{D}_{i}\bar{x}^{T}_{i,j}\parallel_{F}^2 + \lambda\parallel\bar{x}^{T}_{i,j} \parallel\\  
	&\text{with} \; i \in [ 1,N], \; j \in [t^{1}_{i},t^{\mathrm{end}}_{i}]
	\end{split}
	\end{equation}
	At each outer iteration, we update a new $\mu$ and optimize the Eq~(\ref{eq:gm}). The solution obtained at each iteration is used as an initial guess for the next iteration. The final solution is computed until the original non-convexity function is recovered ($\mu = 1$).
	\begin{equation}\label{eq:gm}
	\min_{\forall \mathbf{D}_i \in \mathcal{D}, \forall \bar{x}_{i,j} \in \mathcal{X}}\sum_{j=t^1_{i}}^{t^{\mathrm{end}}_{i}}g_{\mu}\left( e(\bar{y}_{i,j},\mathbf{D}_{i}\bar{x}_{i,j}^{T})\right)
	\end{equation}
	We use the Black-Rangarajan duality to combine the GNC-GM function with weighted ODL cost as follow:
	\begin{multline}\label{eq:GODL}
		\min_{\forall \mathbf{D}_i \in \mathcal{D}, \forall \bar{x}_{i,j} \in \mathcal{X}} \sum_{j=t^1_{i}}^{t^{\mathrm{end}}_{i}} 	\mathcal{O}_{i,j}(w_{i,j},\mathbf{D}_i,\bar{x}_{i,j}^{T}) + \Phi_{g_{\mu}}(w_{i,j})
	\end{multline}
	with weighted cost:
	\begin{equation}\label{eq:odl}
	\begin{split}
	\mathcal{O}_{i,j}=&w^{2}_{i,j}\left( \frac{1}{2}\parallel \bar{y}_{i,j}-\mathbf{D}_{i}\bar{x}^{T}_{i,j}\parallel_{F}^2 + \lambda\parallel\bar{x}^{T}_{i,j}\parallel\right)  \\
	=&\frac{1}{2}\parallel w_{i,j}\bar{y}_{i,j}-\mathbf{D}_{i}(w_{i,j}\bar{x}^{T}_{i,j})\parallel_{F}^2 + \\
	&\lambda w_{i,j}\parallel w_{i,j}\bar{x}^{T}_{i,j}\parallel
	\end{split}
	\end{equation}
	and penalty term:
	\begin{equation}\label{eq:penalty}
	\Phi_{g_{\mu}} = \mu_{i}c^2(w_{i,j}-1)^2
	\end{equation}
	With simplified expression of $\bar{\hat{x}}^T = w\bar{x}^T$, $\bar{\hat{y}}^T = w\bar{x}^T$ and  $\lambda_w = w\lambda$, the Eq~(\ref{eq:odl}) can be described as following:
	\begin{equation}\label{eq:odl2}
	\mathcal{O}_{i,j}=\frac{1}{2}\parallel \bar{\hat{y}}_{i,j}-\mathbf{D}_{i}\bar{\hat{x}}^{T}_{i,j}\parallel_{F}^2 + \lambda_{w\;i,j}\parallel \bar{\hat{x}}^{T}_{i,j}\parallel
	\end{equation}
	At the first inner iteration, all weights are set to $1$. During inner iterating, the weighted ODL is optimized with fixed weight ($w_{i,j}$), and then we optimize over $w_{i,j}$ with a fixed cost of ODL. At a particular inner iteration $k$ within weighted sub-sequence $\mathbf{\hat{Y}}_i$, we perform the following:
	
	1) \textbf{Dictionary Learning}: minimize the Eq~(\ref{eq:GODL}) with respect to $\mathbf{D}_i^{(k)}$ and $\bar{x}_{i,j}^{(k)}$ with fixed $w_{i,j}^{(k-1)}$. This problem is the original ODL, but with weighted training sequence:
	\begin{equation}
	\min_{\forall \mathbf{D}_i \in \mathcal{D}, \forall \mathbf{X}_i \in \mathcal{X}} \sum_{i=1}^{N}\frac{1}{2}\parallel\mathbf{\hat{Y}}_{i}-\mathbf{D}_{i}\mathbf{\hat{X}}_{i} \parallel_{F}^{2} + \lambda_{w\;i}\parallel\mathbf{\hat{X}}_{i}\parallel
	\end{equation}
	In ODL optimization, we first update coefficient matrix $\mathbf{X}_{i}^{(k)}$  with fixed action unit $\mathbf{D}_{i}^{(k-1)}$ (Sparse Coding). We assign the weight parameter to training sequence $\mathbf{Y}_{i}$ and coefficient matrix $\mathbf{X}_{i}^{(k)}$. Then update action unit $\mathbf{D}_{i}^{(k)}$ with fixed weighted coefficient matrix $\mathbf{\hat{X}}_{i}^{(k)}$ and weighted input matrix  $\mathbf{\hat{Y}}_{i}$ (Dictionary Learning):
	\begin{itemize}
		\item Assign Weight: $\mathbf{\hat{Y}}_{i} = \bar{w}_{i}^{T}\cdot\mathbf{Y}_{i}$ and $\bar{\lambda}_w = \bar{w}_{i}^{T}\lambda$, where $\cdot$ is column dot-production.
		\item Sparse Coding: we use Lasso-Fista algorithm to update $\mathbf{\hat{X}}_{i}^{(k)}$ with fixed $\mathbf{D}_{i}^{(k-1)}$, see \cite{mairal2010online}.
		\item Dictionary Learning: minimize the following equation with fixed $\mathbf{\hat{X}}_{i}^{(k)}$:
		\begin{equation}
		\begin{split}
		\mathbf{D}_{i}^{(k)}=&\argmin_
		{\forall \mathbf{D}_i \in \mathcal{D}}-2\mathrm{tr}(\mathbf{E}_i^T\mathbf{D}_i^{(k-1)})  \\
		&+\mathrm{tr}(\mathbf{D}_i^{(k-1)}\mathbf{F}_i\mathbf{D}_i^{T(k-1)}) \\
		&\text{with} \;\mathbf{E}_i = \mathbf{\hat{Y}}_i\mathbf{\hat{X}}_i^{T(k)} \\
		&\text{and} \;\mathbf{F}_i = \mathbf{\hat{X}}_i^{(k)}\mathbf{\hat{X}}^{T(k)}_i
		\end{split}
		\end{equation}
	\end{itemize}
	
	2) \textbf{Weight update}: minimize the Eq~(\ref{eq:GODL}) with respect to weight $w_{i,j}^{(k)}$ with fixed dictionary matrix $\mathbf{D}_i^{(k)}$ and coefficient vector $x_{i,j}^{(k)}$.
	\begin{equation}
	\begin{split}
	\bar{w}_{i}^{T(k)} =&\argmin_{w_{i,j} \in [0,1]} \sum_{j=t_i^1}^{t_i^{\mathrm{end}}} \{ 	\mathcal{O}_{i,j}(w_{i,j}^{(k-1)},\mathbf{D}_i^{(k)},\bar{x}_{i,j}^{T(k)})\\ + &\Phi_{g_{\mu}}(w_{i,j}^{(k-1)})\} 
	\end{split} 
	\end{equation}
	Using introduced ODL function Eq~(\ref{eq:odl}) and penalty function Eq~(\ref{eq:penalty}), the weight update at iteration $k$ can be solved in form as:
	\begin{equation}
	w_{i,j}^{2(k)} = \left( \frac{\mu_{k} c^2 }{\mu_{k} c^2+e_{i,j}^{2}}	\right) ^2 
	\end{equation}
	where $e_{i,j}^2$ is Frobenius norm of error between training sequence $\bar{y}_{i,j}$ and approximation model $\mathbf{D}_i\bar{x}_{i,j}$, see Eq~(\ref{eq:error}). 
	
	In the implementation, we start with an initialization $\mu_{0} = 2*e^2_{i,\mathrm{end}}/c^2$ with 
	\begin{equation}
	e^2_{i,\mathrm{end}} \dot{=} \max_{\forall \mathbf{e}_{i,j} \in \mathcal{E}_i,j\in[t_i^0,t_i^{\mathrm{end}}]} e_{i,j}^2
	\end{equation}

	At each outer iteration, update $\mu_{k} = \mu_{k-1}/1.4$ and stop when $\mu_{k}$ is blow $1$, see \cite{yang2020graduated}.
	
	\subsection{Inferring phase}
	In the inferring phase, we assume that the error between sub-sequence $\mathbf{Y}_i$ and model $\mathbf{D}_i\mathbf{X}_i$ is normally distributed. Hence, the measured error $e_i$ between real-time skeleton frames of a fall-down action and the action unit model should fall within the confidence interval as follows:
	\begin{equation}\label{eq:unit_model}
	    \frac{|e_i-e_{i,\text{mean}}|}{\sigma(e_i)} < \alpha
	\end{equation}
	 where $e_{i,\mathrm{mean}}$ is the mean error of training set, $\sigma({e}_i)$ is the standard deviation of error $e_i$ ,and $\alpha$ is an acceptance parameter. 
	 
	 Since the fall event has strict order of sub actions, which are from \textit{"standing"} to \textit{"on the ground"}, each sub-action detection will be performed only when the previous action is passed.
	 
    %

    In addition to action unit extraction, the temporal feature of fall down is important as well. A fall is defined as an event that results in a person moving from a higher to a lower level, typically rapidly and without control. From this definition, we can know that the action fall down is a rapid human’s height change in a very short time. For the height change, we don’t need all the skeleton information. We only need the skeleton information in $y$-direction as the following equation:
    \begin{equation}
        h = y^T_{max}-y^T_{min}
    \end{equation}
    where y is the value of skeleton in y axis, h means the height of skeleton and T is the width of time interval shifted from beginning of video to end. Since the first action unit is "\textit{standing}", we define its height as an initial value $h_{init}$. The height change of fall event inside a time interval should meet following two conditions:
    
    \begin{equation}\label{eq:temporal_condition}
      \begin{cases}
        \frac{h^0}{h_{init}}>0.9, &h^0 \text{ is begin of interval}.\\
        \frac{h^{T-1}}{h^0}<0.5, & h^{T-1} \text{ is end of interval}.
      \end{cases}
    \end{equation}
    where these thresholds are obtained through experiments.
    
	\section{EXPERIMENTS AND RESULTS}
	\label{sec:4}
	In this section, we present the experiments' results on the NTU RGB+D dataset \cite{Shahroudy_2016_NTURGBD}. First, it introduces the dataset for training and evaluation. Second, it displays the tendency of weight parameter with an increasing number of iteration in the training phase and demonstrates how the dimension of action units influences the prediction performance. In the end, we compare our method with other existing well-performed dictionary learning methods on the NTU RGB+D dataset \cite{Shahroudy_2016_NTURGBD}.

	\begin{figure*}[t]
		\vspace{0.13cm}
        \begin{center}
		\addtolength{\tabcolsep}{-5pt}    
		\renewcommand{\arraystretch}{0.1}
		\begin{tabular}{rcc} 
			\includegraphics[width=0.2\linewidth]{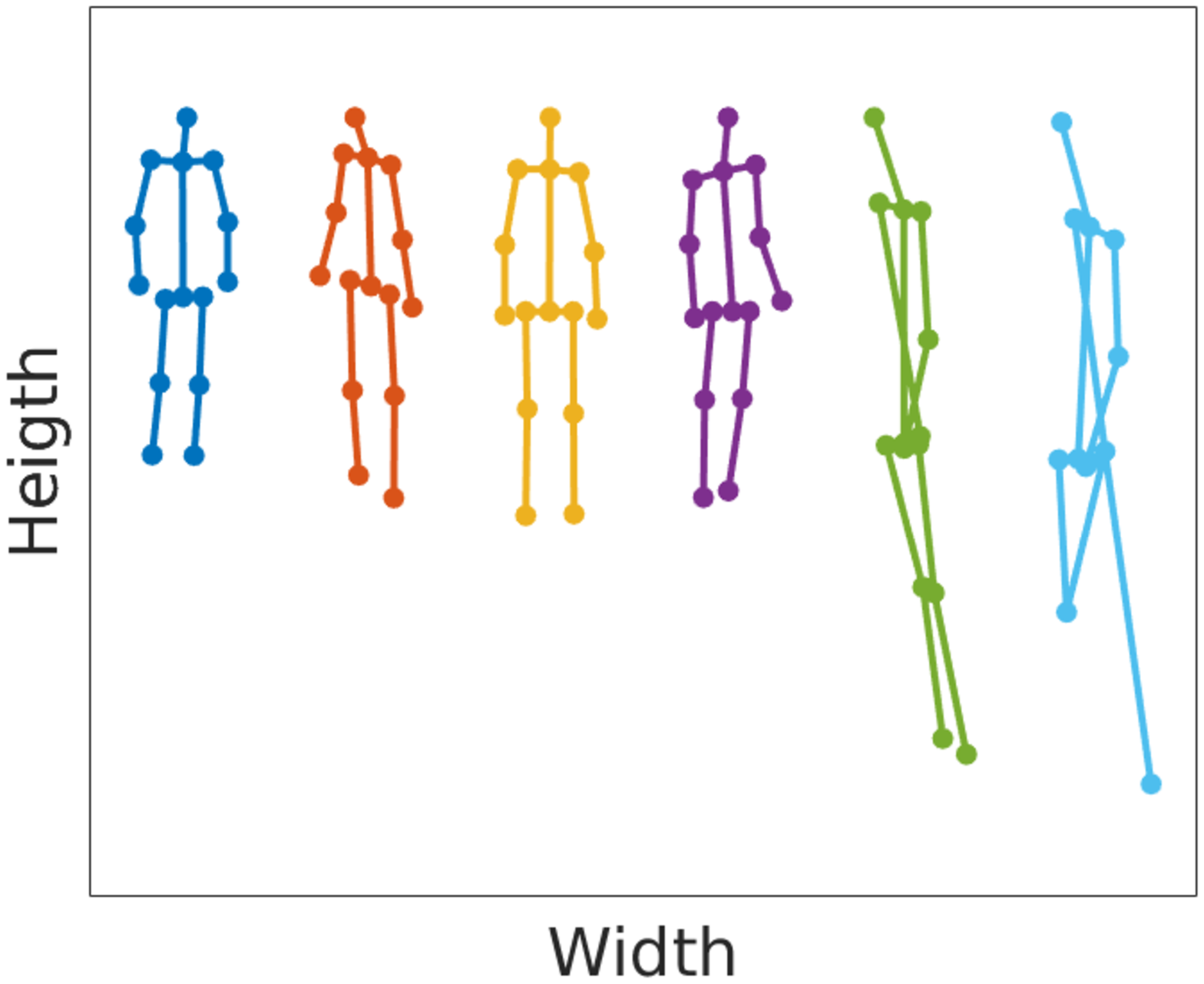} &
			\multirow{2}[2]{*}[25mm]{\includegraphics[width=0.46\linewidth]{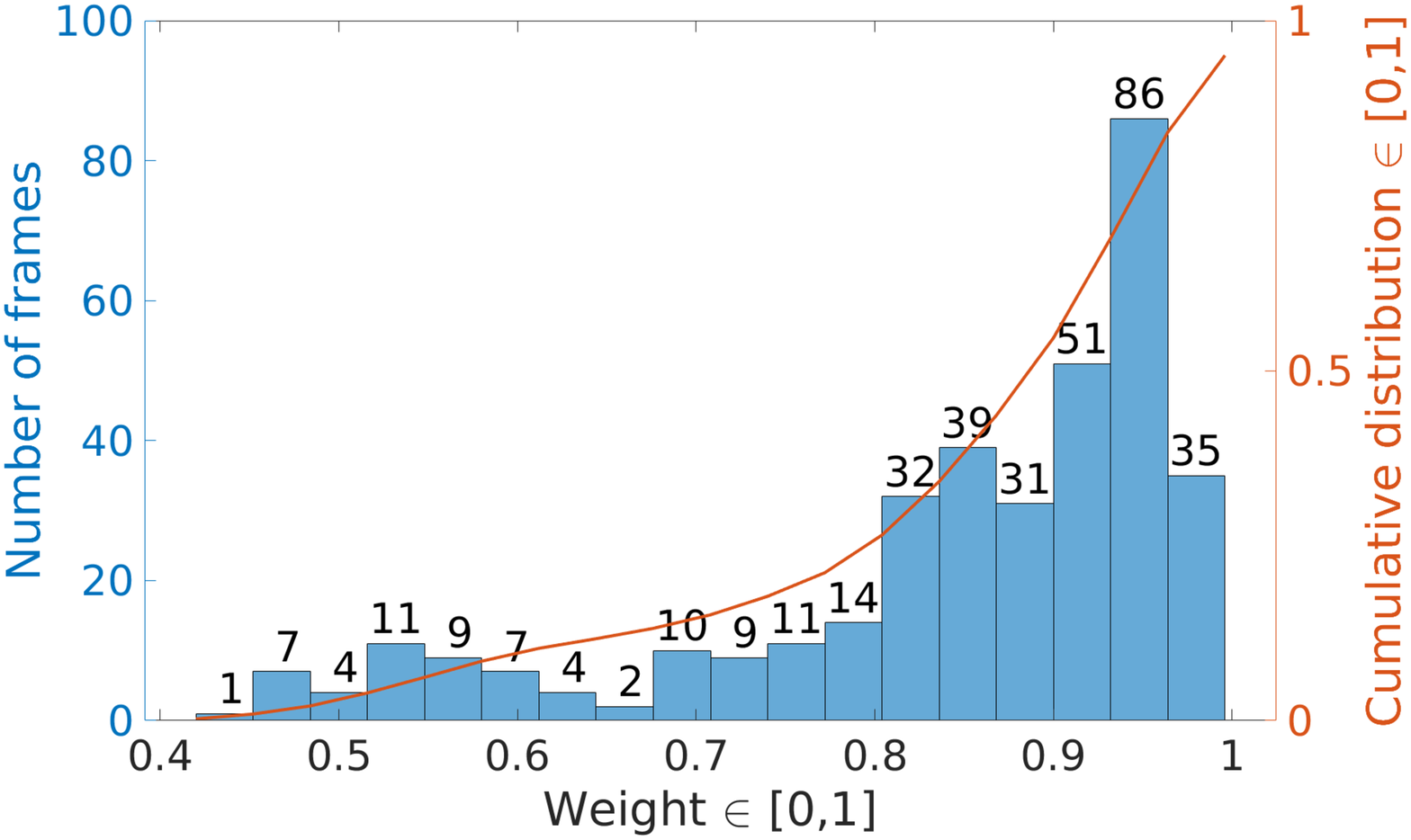}} &
			\multirow{2}[2]{*}[25mm]{\includegraphics[width=0.34\linewidth]{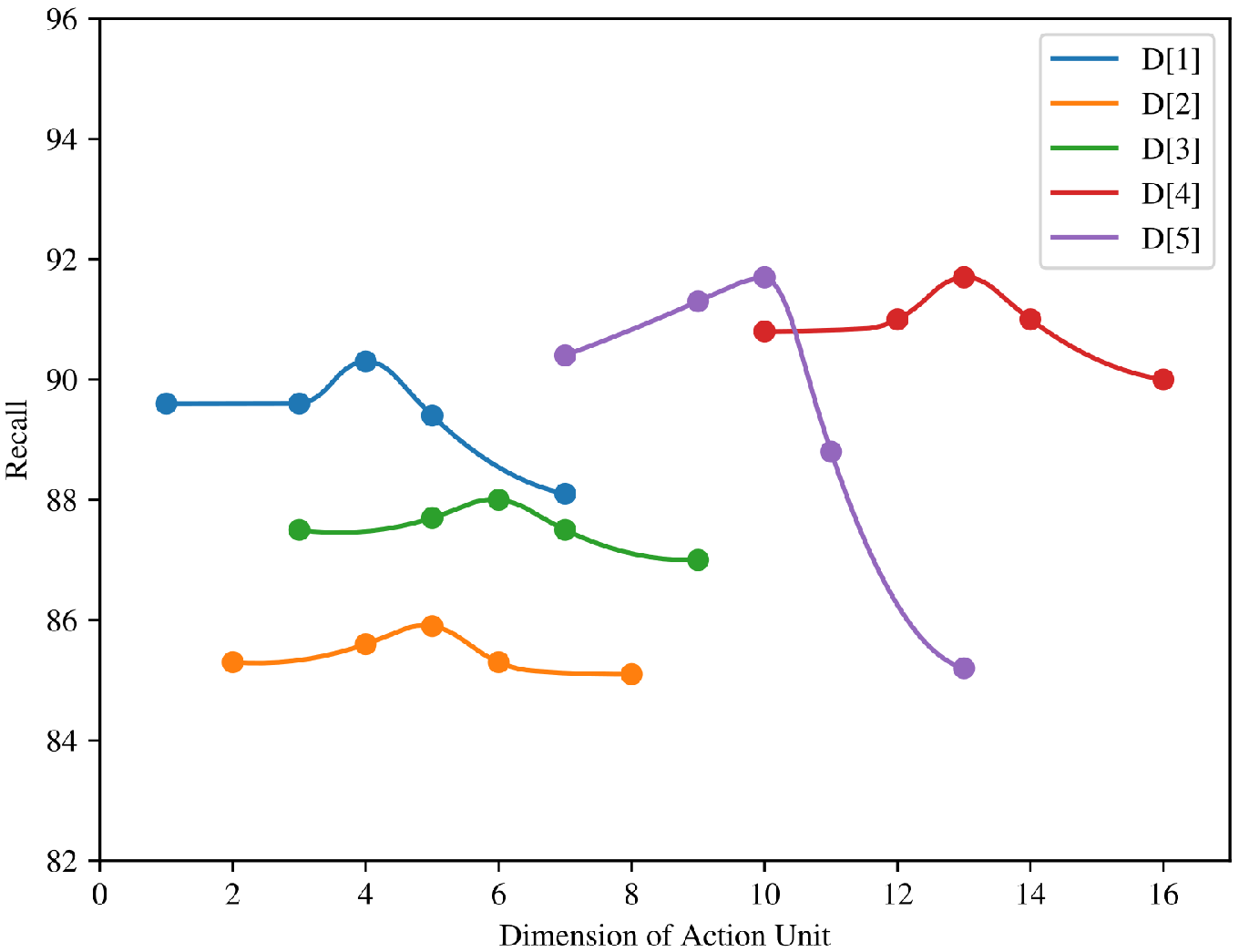}}
			\\ 
			\includegraphics[width=0.2\linewidth]{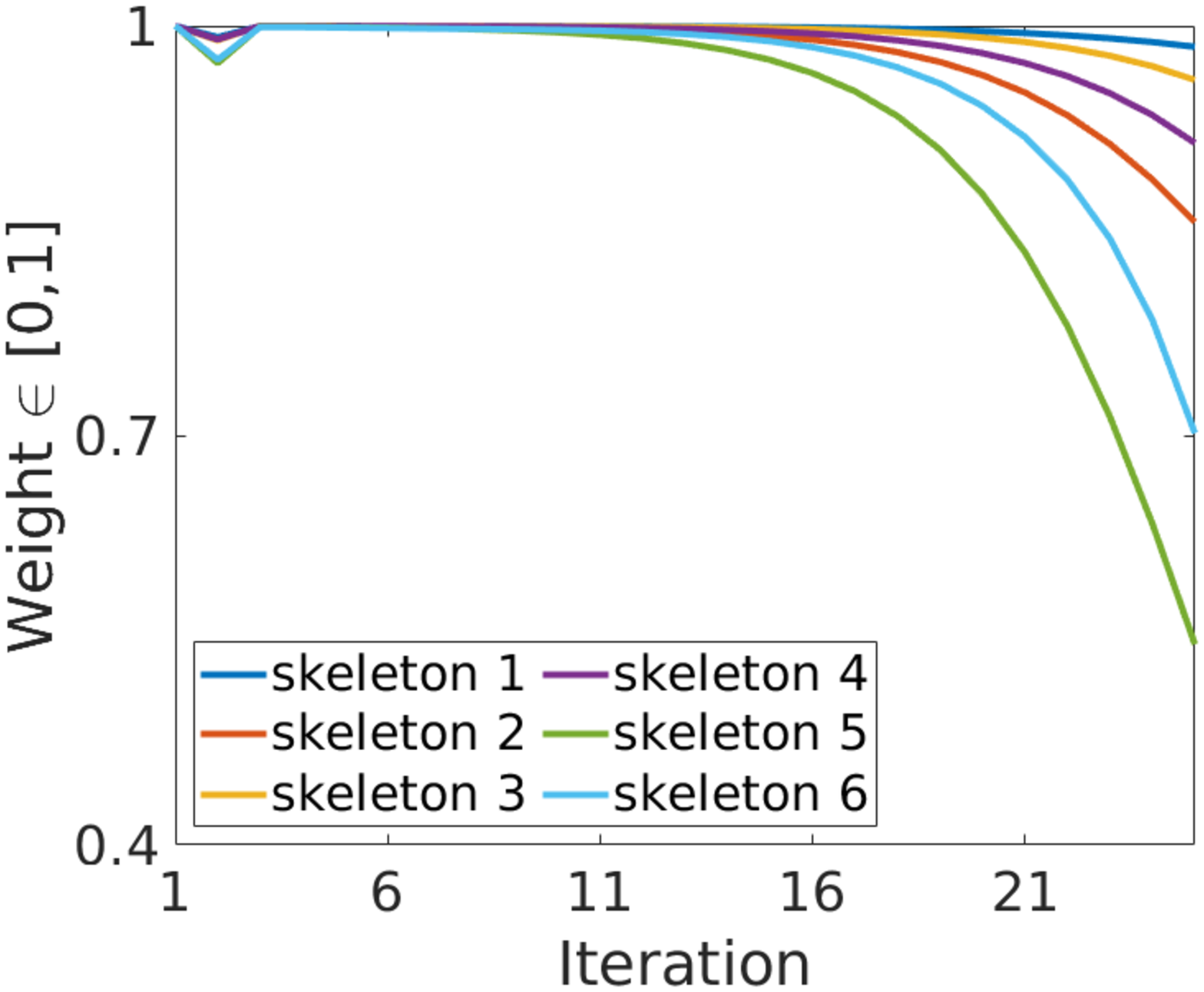}  &  & \vspace{0.15cm}\\
			
			\multicolumn{1}{c}{(a)} & (b) &(c) 
		\end{tabular}
		\caption{(a) The weight tendency of six skeletons in the first action unit "standing" over iteration (b) the histogram of weight value and its cumulative distribution at the last iteration. (c) The recall with different action unit dimension}
		\label{fig:weight}
		\end{center}
	\end{figure*}
    \subsection{Dataset}
    
    \textbf{NTU RGB+D dataset} \cite{Shahroudy_2016_NTURGBD} is one of the largest datasets for human action recognition. It contains $60$ action classes, $56,880$ video samples, and their depth image frames. From the first $9$ setups ($s001 - s009$), we successfully generate $209$ fall-down 3D skeleton examples, in which $105$ samples are used for training and the rest $104$ are used for testing. For both training and testing, we randomly select subjects and camera views, including two side views, two diagonal views and a front view

	In order to recognize fall-down event from similar actions, we generate $420$ sitting-down and $405$ ground-lift 3D skeleton examples and merge them into the test dataset, the rest $280$ skeleton examples are taken from $46$ other actions.

    As suggested in the dataset \cite{Shahroudy_2016_NTURGBD}, we use cross-subject (CS) and cross-view (CV) criteria to compare our model with deep learning method. In CS evaluation, the subjects are split into training and testing groups. The IDs of training subjects are 2,3,5,7,8. The result of CS evaluation is shown in table \ref{tab:1}. In CV evaluation, the samples of camera view 2 and 3 are used for training. Camera view 2 and 3 include one front view and two side views.
    The samples of camera view 3 are used for testing, which contains diagonal views.


	\subsection{Validating the effectiveness of GODL}
	
    
	In order to prove that GODL is resistant to outliers, we record $6$ skeleton's weight $w$ change during training of the first action unit, see Fig~\ref{fig:weight} (a). It shows the weight tendency of six skeleton examples in the first sub-sequence "\textit{standing}" over iteration in the GODL program. The weight of outliers (skeleton $5$ and $6$) have a steeper decreasing trend, while the inlier's (skeleton $1$ - $4$) weight changes slower. At the end of the iteration, outliers are assigned with $0.547$ and $0.717$, respectively. In opposite to outliers, the weight of inliers still keeps a high value, respectively $0.9847$, $0.8564$, $0.9604$, and $0.9142$. Fig~\ref{fig:weight} (b) shows the histogram of weight and its cumulative distribution at the last iteration. Fast $90\%$ of skeletons have a weight with a value larger than $0.6$. These skeletons have a greater impact on the cost function. Hence these $90\%$ skeletons are considered as inliers, and the last $10\%$ with lower value are outliers.

    Since each action unit's dimension influences the performance of prediction, we measure its performance of recall with $6$ different settings and select the optimal dimension for each action unit. 
	The dimension of dictionary is highly depending on the complexity of action unit, for example the first three dictionary $\mathbf{D}_1$, $\mathbf{D}_2$ and $\mathbf{D}_3$ has less dimension than the last two dictionary $\mathbf{D}_4$ and $\mathbf{D}_5$, because action unit "\textit{standing}" "\textit{bending knee}" and "\textit{opening arm}" are much simple than "\textit{knee landing}" and "\textit{arm supporting}". Before reaching the optimal point, the recall increases with dimension, because it is not enough to represent the action space. After exceeding the optimal point, the recall is decreasing with dimension because of overfitting. Fig~\ref{fig:weight} (c) presents the selection process of each unit dimension. The optimal combination of dimensions is $\{4,5,6,10,13\}$.
	
		
	
	\begin{figure*}[t]
		\vspace{0.13cm}
        \begin{center}
		\addtolength{\tabcolsep}{-5pt}    
		\renewcommand{\arraystretch}{0.1}
		\begin{tabular}{cc} 
            \includegraphics[width=0.66\linewidth]{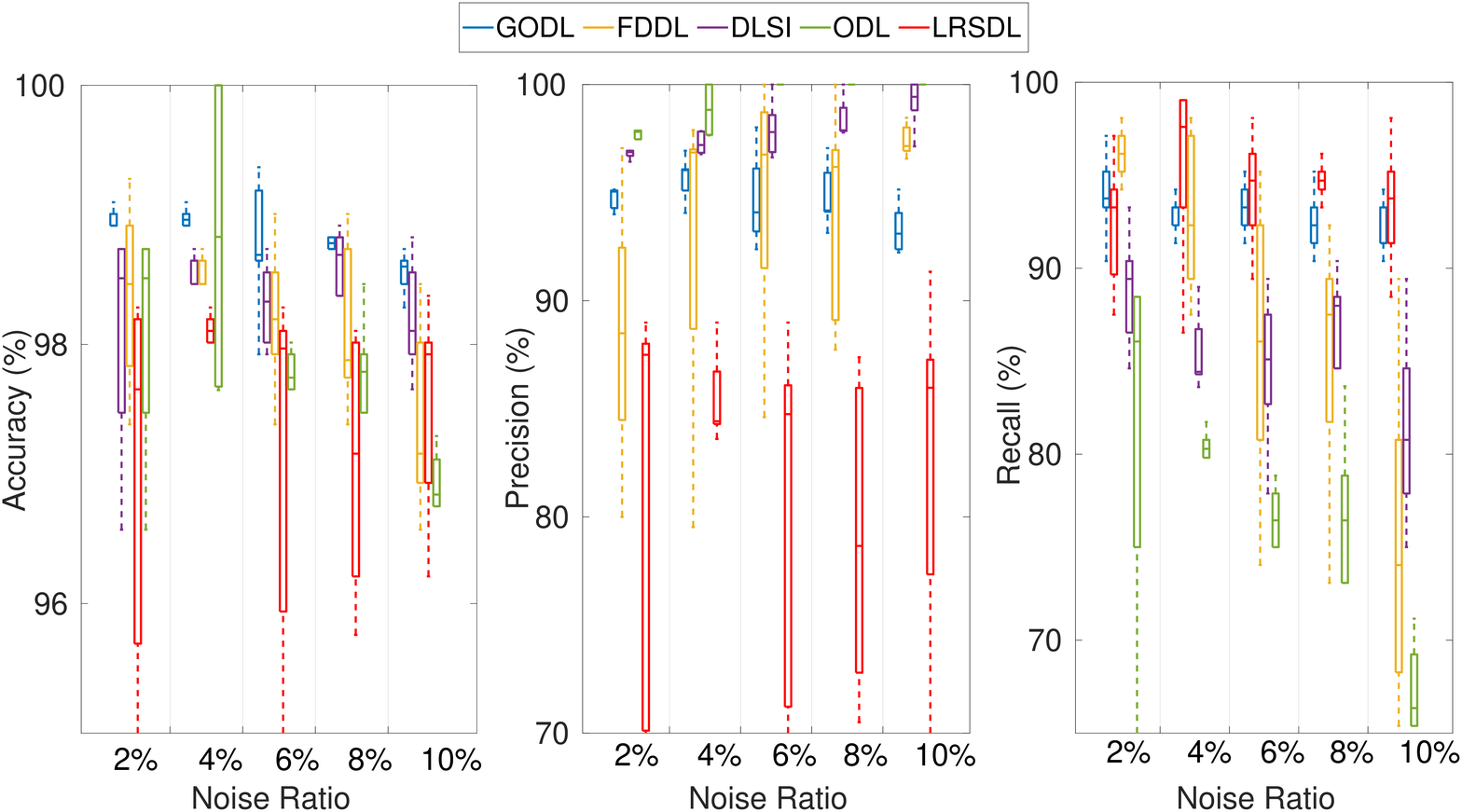} &
            \includegraphics[width=0.32\linewidth]{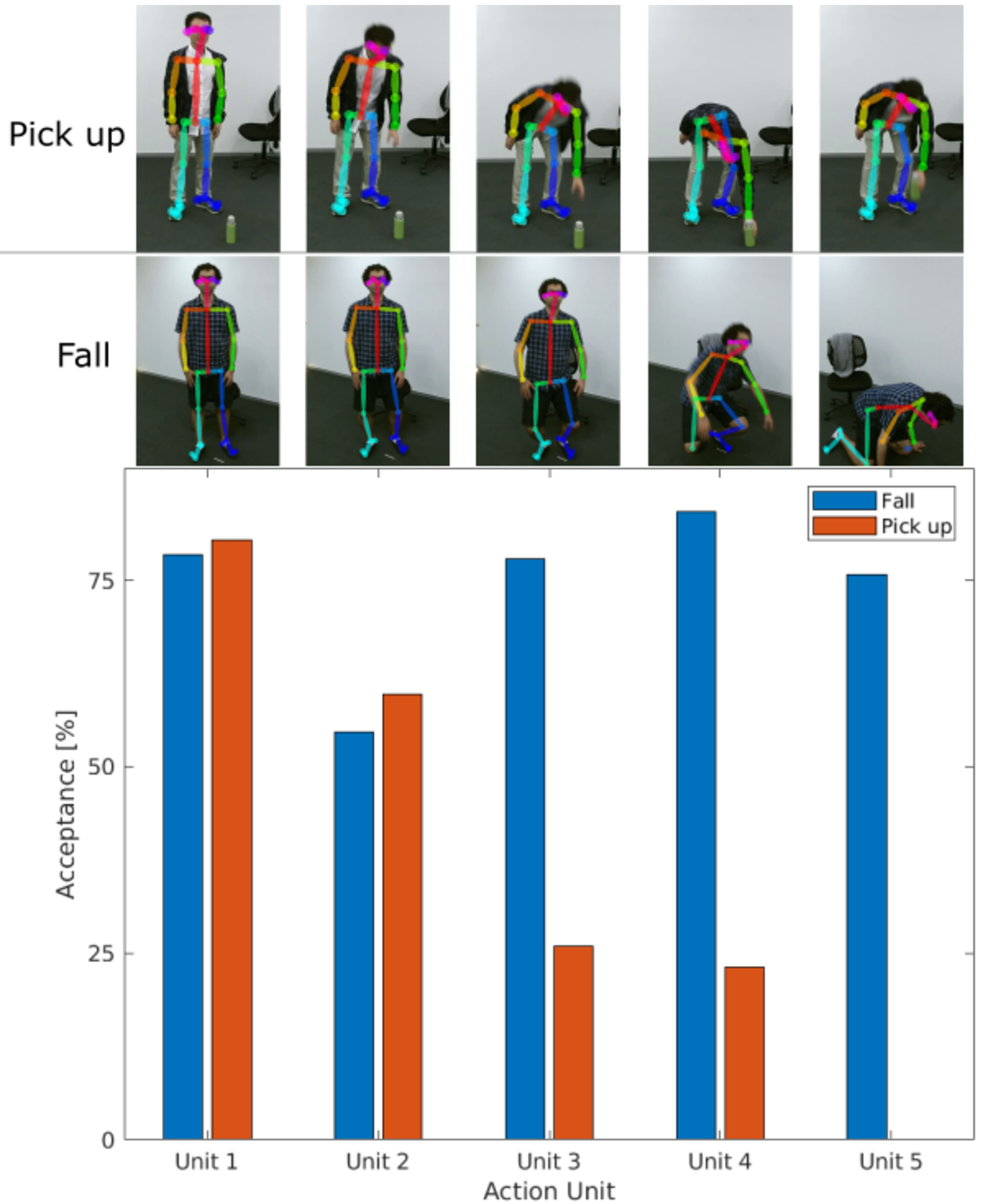} \\
	         \vspace{0.15cm}\\
	         (a) & (b)
		\end{tabular}
		\caption{(a) The comparison of robustness under different noise level. (b) Visualization of prediction process }
		\label{fig:noise}
		\end{center}
	\end{figure*}
	
    \subsection{Evaluation of fall-down using action unit and temporal structure}

	\begin{table}[t]
		\centering
		\caption{Performance Comparison with existing Dictionary Learning methods}
		\resizebox{\linewidth}{!}{
			\begin{tabular}{|l|l|l|l|}
				\hline
				& Accuracy (\%)& Recall (\%) & Precision (\%) \\ \hline
				ODL \cite{mairal2010online} &$ 98.86\pm 0.29 $ & $92.40\pm \mathbf{1.63}$ & $95.36\pm 2.37 $ \\ 
				DLSI \cite{ramirez2010classification} &$ 98.71\pm \mathbf{0.28} $ & $92.21\pm 2.79$ & $94.01\pm 2.77 $ \\
				FDDL \cite{yang2011fisher} &$ 98.25\pm 1.34 $ & $95.77 \pm 3.65$ & $87.23\pm 5.75 $ \\
				LRSDL \cite{vu2016learning}&$ 98.11\pm 0.75 $ & $\mathbf{96.79}\pm 2.10$ & $85.29\pm 4.15 $ \\
				\hline
				GODL (our)  & $\mathbf{99.00}\pm 0.36$  & $94.23\pm 2.6$ & $\mathbf{95.62}\pm\mathbf{1.53} $  \\ \hline
		\end{tabular}}
		\label{tab:1}
		\begin{tablenotes}
            \item The best results of each class are in \textbf{bold}.
        \end{tablenotes}
	\end{table}

	The evaluation results can be found in Tab~\ref{tab:1}. Compared to the other four state-of-the-art Dictionary Learning methods, our GODL model achieves the best accuracy and precision. Considering the recall, FDDL \cite{yang2011fisher} and LRSDL \cite{vu2016learning} both achieves a good performance. However, their precision is $8\% \sim 10\%$ lower than our method. Compared to the baseline ODL \cite{mairal2010online}, our method has better performance in all aspects.

	To demonstrate the robustness of our method, we deliberately add noise into the training data and compare the performance of our method with other methods. Fig~\ref{fig:noise}(a) shows the accuracy, precision and recall of the methods under different noise ratio ($2\% \sim 10\%$). Although some methods surpass our methods in recall and precision, overall, with the noise increases, our method remains at the same level and the highest in the accuracy. It proves that our method is more robust than the other four methods.

	In addition to the dictionary learning methods, we also compare our result with the state-of-the-art deep learning based fall-down detection methods. The comparison is shown in table \ref{tab:3}. The deep learning based methods have slightly higher precision.
	However, our method has more stable prediction result. Besides, our method encodes spatial-temporal information, which is more explainable than the end-to-end deep learning method. It is not possible in the end-to-end deep learning methods, because they can only detect the fall-down action after falling. Compared to other spatial-temporal methods, our method can determine which step of the fall-down action is in progress, see Fig \ref{fig:noise} (b). For more real-time evaluation, please check the attached video material.

	\begin{table}[t]
		\centering
		\caption{Performance Comparison with existing Deep Learning methods (CV + CS)}
		\resizebox{\linewidth}{!}{
			\begin{tabular}{|l|l|l|}
				\hline
				& Accuracy (\% CS) & Accuracy (\% CV) \\ \hline
				ST-GCN \cite{ST-GCN}&$ 97.03\pm 0.83 $ & $97.45\pm 1.11$\\
				Biomechanic, RNN \cite{Xu_Zhou_2018}&$ 97.40\pm 1.25 $ & $97.20\pm 1.79$\\
				Thining, DNN \cite{Thinning_DNN}&$ \textbf{99.20}\pm 1.10 $ & $\textbf{99.20}\pm 1.56$\\
				\hline
				GODL (our) & $98.41\pm \textbf{0.04}$ & $99.03 \pm \textbf{0.15} $  \\ \hline
		\end{tabular}}
		\label{tab:3}
		\begin{tablenotes}
            \item The best results of each class are in \textbf{bold}.
        \end{tablenotes}
	\end{table}

	\section{CONCLUSIONS}
	\label{sec:5}
	In the paper, we have proposed a novel event detection method using robust latent action units extraction method GODL and performed at the example of fall-down detection. Experiments have been evaluated on a public dataset. The proposed method outperforms the existing good dictionary learning methods on both robustness and average accuracy.

	Compared to the end-to-end deep learning methods, our method includes spatial-temporal information, which is better explainable. Compared to other spatial-temporal methods, our method can determine which step of the fall-down action is in progress, instead of determining a single fall-down action. In other words, our method contains implicit information. Therefore, our method can not only detect fall-down activity, but also predict and prevent the fall-down activity. It is very useful in scenarios such as health-care area. Most importantly, we attempt to approximate action space through more mathematical method.
	
	We plan to focus on applying the proposed method to recognize different actions with larger datasets in the future. 
	
	\bigskip
	
	\section*{ACKNOWLEDGMENT}
	We gratefully acknowledge the funding of the Lighthouse Initiative Geriatronics by StMWi Bayern (Project X, grant no. 5140951) and LongLeif GaPa GmbH (Project Y, grant no. 5140953).

	



	
	
	\bibliographystyle{IEEEtran}
	\bibliography{IEEEexample}

\end{document}